\documentclass[10pt,twocolumn,letterpaper]{article}

\usepackage{iccv}
\usepackage{times}
\usepackage{epsfig}
\usepackage{graphicx}
\usepackage{amsmath}
\usepackage{amssymb}
\usepackage{booktabs}
\usepackage{multirow}
\usepackage{caption}
\usepackage{subcaption}
\usepackage[table,xcdraw]{xcolor}

\def\mI{{\mathcal I}}

\def\0{{\bf 0}}
\def\1{{\bf 1}}

\def\bA{{\bf A}}

\def\bE{{\bf E}}

\def\bK{{\bf K}}

\def\bO{{\bf O}}

\def\bQ{{\bf Q}}

\def\bV{{\bf V}}
\def\bW{{\bf W}}
\def\bX{{\bf X}}

\def\bx{{\bf x}}
\def\by{{\bf y}}

\newlength\savewidth\newcommand\shline{\noalign{\global\savewidth\arrayrulewidth
  \global\arrayrulewidth 1pt}\hline\noalign{\global\arrayrulewidth\savewidth}}

\usepackage[pagebackref=true,breaklinks=true,letterpaper=true,colorlinks,bookmarks=false]{hyperref}

\iccvfinalcopy 

\ificcvfinal\fi

\definecolor{mypink}{rgb}{0.858, 0.188, 0.478}

\usepackage{xcolor}
\usepackage{hyperref}
\hypersetup{
  colorlinks,
  citecolor=violet,
  linkcolor=red
}

 \makeatletter
\def\@fnsymbol#1{\ensuremath{\ifcase#1\or \dagger\or \ddagger\or
   \mathsection\or \mathparagraph\or \|\or **\or \dagger\dagger
   \or \ddagger\ddagger \else\@ctrerr\fi}}
\makeatother

\begin{document}

\title{Scalable Vision Transformers with Hierarchical Pooling}

\author{
Zizheng~Pan
\quad Bohan~Zhuang\thanks{
Corresponding author. Email: $\tt bohan.zhuang@monash.edu$}
\quad Jing~Liu 
\quad Haoyu~He 
\quad Jianfei~Cai \\[0.12cm]
{Dept of Data Science and AI, Monash University} 
}

\maketitle

\ificcvfinal\thispagestyle{empty}\fi

\begin{abstract}
The recently proposed Visual image Transformers (ViT) with pure attention have achieved promising performance on image recognition tasks, such as image classification.
However, the routine of the current ViT model is to maintain a full-length patch sequence during inference, which is redundant and lacks hierarchical representation. To this end, we propose a Hierarchical Visual Transformer (HVT) which progressively pools visual tokens to shrink the sequence length and hence reduces the computational cost, analogous to the feature maps downsampling in Convolutional Neural Networks (CNNs). It brings a great benefit that we can increase the model capacity by scaling dimensions of depth/width/resolution/patch size without introducing extra computational complexity due to the reduced sequence length. Moreover, we empirically find that the average pooled visual tokens contain more discriminative information than the single class token. To demonstrate the improved scalability of our HVT, we conduct extensive experiments on the image classification task. With comparable FLOPs, our HVT outperforms the competitive baselines on ImageNet and CIFAR-100 datasets. Code is available at \url{https://github.com/MonashAI/HVT}.
\end{abstract}

\section{Introduction}

\begin{figure}[!t]
	\centering
	\includegraphics[width=\linewidth]{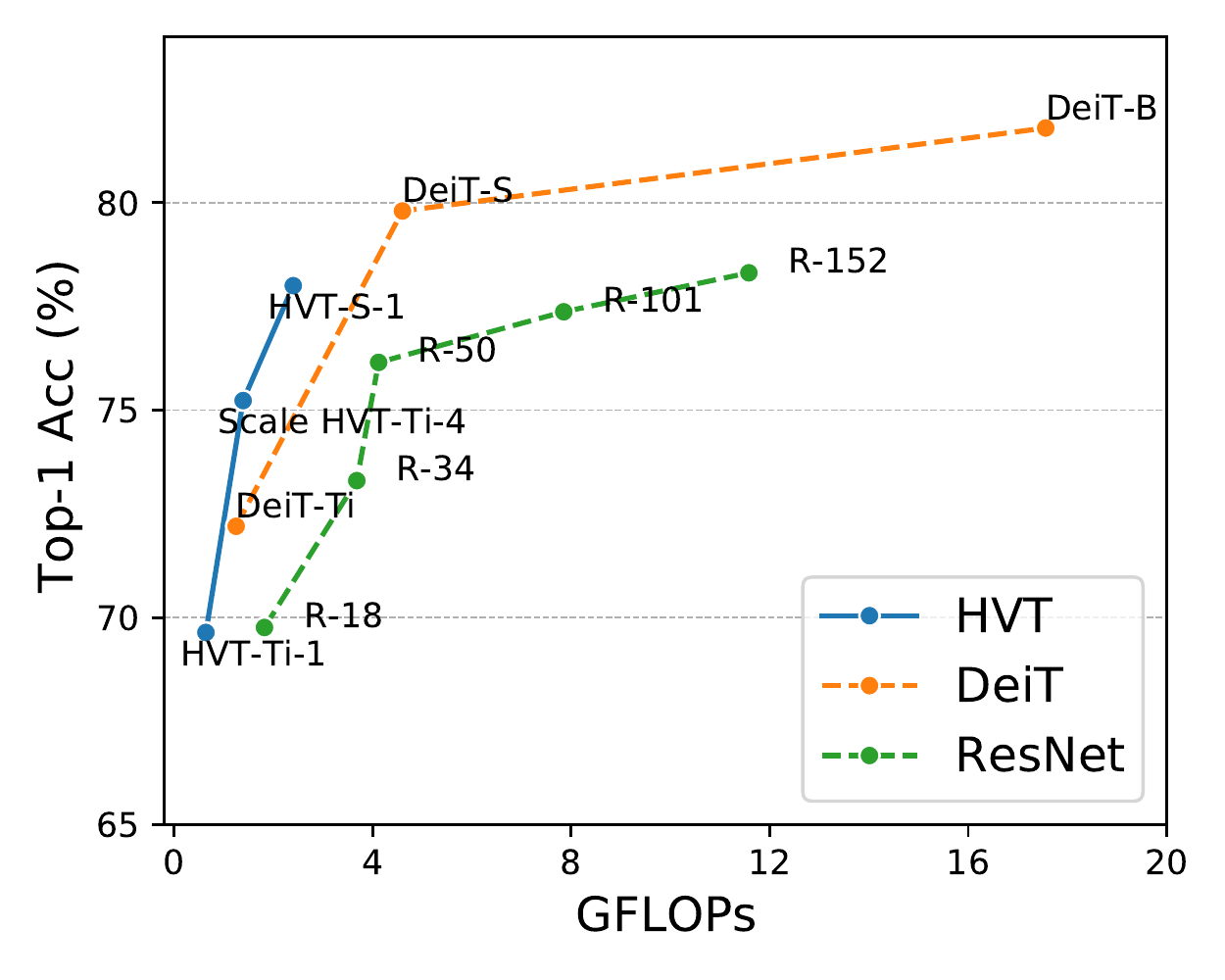}
	\caption{Performance comparisons on ImageNet. With comparable GFLOPs (1.25 vs. 1.39), our proposed Scale HVT-Ti-4 surpasses DeiT-Ti by 3.03\% in Top-1 accuracy.
	}
	\label{fig:flops-acc}
	\vspace{-10pt}
\end{figure}

Equipped with the self-attention mechanism that has strong capability of capturing long-range dependencies, Transformer \cite{transformer} based models have achieved significant breakthroughs in many computer vision (CV) and natural language processing (NLP) tasks, such as machine translation \cite{Devlin2019BERTPO, dai2019transformer}, image classification \cite{vit, deit}, segmentation \cite{vistr, maxdlab} and object detection \cite{detr, dedetr}. However, the good performance of Transformers comes at a high computational cost. For example, a single Transformer model requires more than 10G Mult-Adds to translate a sentence of only 30 words. Such a huge computational complexity hinders the widespread adoption of Transformers, especially on resource-constrained devices, such as smart phones.

To improve the efficiency, there are emerging efforts to design efficient and scalable Transformers. On the one hand, some methods follow the idea of model compression to reduce the number of parameters and computational overhead. Typical methods include knowledge distillation \cite{jiao2019tinybert}, low-bit quantization \cite{fqt} and pruning \cite{cbert}. On the other hand, the self-attention mechanism has quadratic memory and computational complexity, which is the key efficiency bottleneck of Transformer models. The dominant solutions include kernelization \cite{katharopoulos2020transformers, peng2021random}, low-rank decomposition \cite{linformer}, memory \cite{rae2020compressive}, sparsity \cite{sparse_attn} mechanisms, etc. 

Despite much effort has been made, there still lacks specific efficient designs for Visual Transformers considering taking advantage of characteristics of visual patterns. In particular, ViT models maintain a full-length sequence in the forward pass across all layers. Such a design can suffer from two limitations. Firstly, different layers should have different redundancy and contribute differently to the accuracy and efficiency of the network. This statement can be supported by existing compression methods \cite{tan2019efficientnet, li2017pruning}, where each layer has its optimal spatial resolution, width and bit-width. As a result, the full-length sequence may contain huge redundancy.
Secondly, it lacks multi-level hierarchical representations, which is well known to be essential for the success of image recognition tasks. 

To solve the above limitations, we propose to gradually downsample the sequence length as the model goes deeper. Specifically, inspired by the design of VGG-style \cite{simonyan2014very} and ResNet-style \cite{resnet} networks, we partition the ViT blocks into several stages and apply the pooling operation (\eg, average/max pooling) in each stage to shrink the sequence length. Such a hierarchical design is reasonable since a recent study \cite{selfconv} shows that a multi-head self-attention layer with a sufficient number of heads can express any convolution layers. Moreover, the sequence of visual tokens in ViT can be analogous to the flattened feature maps of CNNs along the spatial dimension, where the embedding of each token can be seen as feature channels.  Hence, our design shares similarities with the spatial downsampling of feature maps in CNNs.
To be emphasized, the proposed hierarchical pooling has several advantages. (1) It brings considerable computational savings and improves the scalability of current ViT models. With comparable floating-point operations (FLOPs), we can scale up our HVT by expanding the dimensions of width/depth/resolution. In addition, the reduced sequential resolution also empowers the partition of the input image into smaller patch sizes for high-resolution representations, which is needed for low-level vision and dense prediction tasks. 
(2) It naturally leads to the generic pyramidal hierarchy, similar to the feature pyramid network (FPN)~\cite{lin2017feature}, which extracts the essential multi-scale hidden representations for many image recognition tasks.

In addition to hierarchical pooling, we further propose to perform predictions without the class token. Inherited from NLP, conventional ViT models \cite{vit, deit} equip with a trainable class token, which is appended to the input patch tokens, then refined by the self-attention layers, and is finally used for prediction. However, we argue that it is not necessary to rely on the extra class token for image classification. To this end, we instead directly apply average pooling over patch tokens and use the resultant vector for prediction, which achieves improved performance. 
We are aware of a concurrent work~\cite{peg} that also observes the similar phenomenon.

Our contributions can be summarized as follows:
\begin{itemize}

\item We propose a hierarchical pooling regime that gradually reduces the sequence length as the layer goes deeper, which significantly improves the scalability and the pyramidal feature hierarchy of Visual Transformers. The saved FLOPs can be utilized to improve the model capacity and hence the performance.

\item Empirically, we observe that the average pooled visual tokens contain richer discriminative patterns than the class token for classification. 

\item Extensive experiments show that, with comparable FLOPs, our HVT outperforms the competitive baseline DeiT on image classification benchmarks, including ImageNet and CIFAR-100. 

\end{itemize}

\section{Related Work}
\paragraph{Visual Transformers.} 
The powerful multi-head self-attention mechanism has motivated the studies of applying Transformers on a variety of CV tasks. 
In general, current Visual Transformers can be mainly divided into two categories.
The first category seeks to combine convolution with self-attention. 
For example, Carion~\etal~\cite{detr} propose DETR for object detection, which firstly extracts visual features with CNN backbone, 
followed by the feature refinement with Transformer blocks.
BotNet~\cite{botnet} is a recent study that replaces the convolution layers with multiheaded self-attention layers at the last stage of ResNet. Other works~\cite{dedetr,ccnet} also present promising results with this hybrid architecture.
The second category aims to design a pure attention-based architecture without convolutions.
Recently, Ramachandran~\etal~\cite{sasa} propose a model which replaces all instances of spatial convolutions with a form of self-attention applied to ResNet.
Hu~\etal~\cite{lrnet} propose LR-Net~\cite{lrnet} that
replaces convolution layers with local relation layers, which adaptively determines aggregation weights based on the compositional relationship of local pixel pairs. Axial-DeepLab~\cite{axiallab} is also proposed to use Axial-Attention~\cite{axial_attn}, a generalization form of self-attention, for Panoptic Segmentation. 
Dosovitskiy~\etal~\cite{vit} 
first transfers Transformer to image classification.
The model inherits a similar architecture from standard Transformer in NLP and achieves promising results on ImageNet, whereas it suffers from prohibitively expensive training complexity.
To solve this, 
the following work
DeiT~\cite{deit} propose a more advanced optimization strategy and a distillation token, with
improved accuracy and training efficiency.
Moreover, T2T-ViT~\cite{t2tformer} aims to overcome the limitations of simple tokenization of input images in ViT and propose to progressively structurize the image to tokens to capture rich local structural patterns.
Nevertheless, the previous literature all assumes the same architecture to the NLP task, without the adaptation to the image recognition tasks. In this paper, we propose several simple yet effective modifications to improve the scalability of current ViT models.
\begin{figure*}[htp!]
	\centering
	\includegraphics[width=\linewidth]{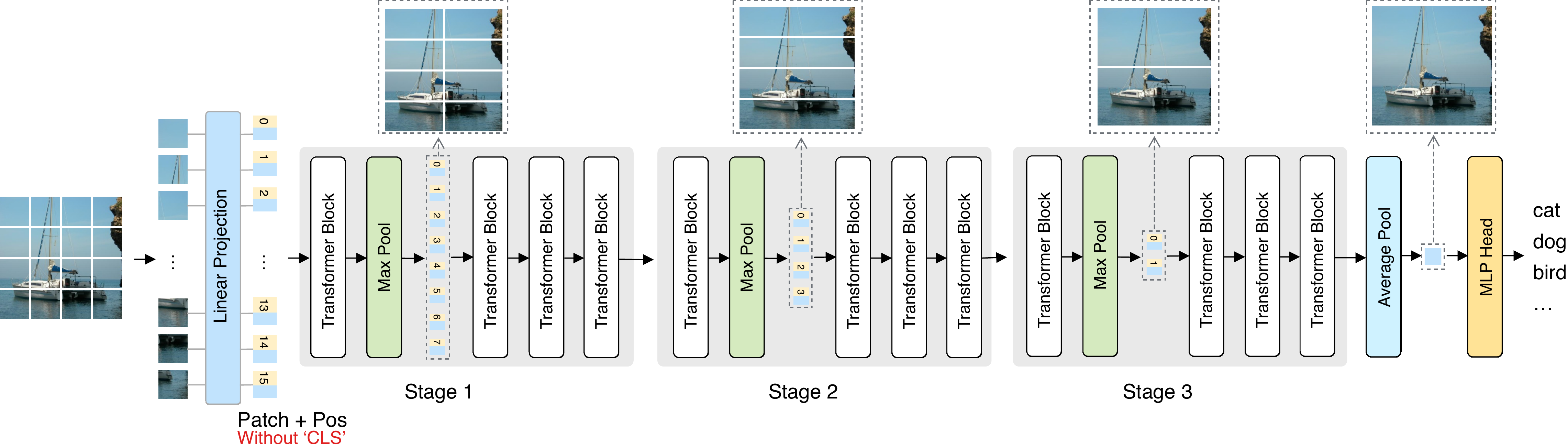}
	\caption{Overview of the proposed Hierarchical Visual Transformer. To reduce the redundancy in the full-length patch sequence and construct a hierarchical representation, we propose to progressively pool visual tokens to shrink the sequence length. To this end, we partition the ViT~\cite{vit} blocks into several stages. At each stage, we insert a pooling layer after the first Transformer block to perform down-sampling. 
	In addition to the pooling layer, we perform predictions using the resultant vector of average pooling the output visual tokens of the last stage instead of the class token only.
	}
	\vspace{3pt}
	\label{fig:model_arch}
\end{figure*}

\paragraph{Efficient Transformers.} Transformer-based models are resource-hungry and compute-intensive despite their state-of-the-art performance. We roughly summarize the efficient Transformers into two categories. The first category focuses on applying generic compression techniques to speed up the inference, either based on quantization \cite{ternarybert}, pruning \cite{sixteen_heads, cbert}, and distillation \cite{distillbert} or seeking to use Neural Architecture Search (NAS) \cite{hat} to explore better configurations. Another category aims to solve the quadratic complexity issue of the self-attention mechanism. 
A representative approach \cite{performer, katharopoulos2020transformers} is to express the self-attention weights as a linear dot-product of kernel functions and make use of the associative property of matrix products to reduce the overall self-attention complexity from $\mathcal{O}(n^2)$ to $\mathcal{O}(n)$. Moreover, some works alternatively study diverse sparse patterns of self-attention \cite{sparse_attn, reformer}, or consider the low-rank structure of the attention matrix \cite{linformer}, leading to linear time and memory complexity with respect to the sequence length. There are also some NLP literatures that tend to reduce the sequence length during processing. For example, Goyal~\etal~\cite{powerbert} propose PoWER-BERT, which progressively eliminates word tokens during the forward pass. Funnel-Transformer ~\cite{funnel} presents a pool-query-only strategy, pooling the query vector within each self-attention layer. 
However, there are few literatures targeting improving the efficiency of the ViT models. 

To compromise FLOPs, current ViT models divide the input image into coarse patches (\ie, large patch size), hindering their generalization to dense predictions. 
In order to bridge this gap, we propose a general hierarchical pooling strategy that significantly reduces the computational cost while enhancing the scalability of important dimensions of the ViT architectures, \ie, depth, width, resolution and patch size. Moreover, our generic encoder also inherits the pyramidal feature hierarchy from classic CNNs, potentially benefiting many downstream recognition tasks. 
Also note that different from a concurrent work~\cite{pvt} which applies 2D patch merging, this paper introduces the feature hierarchy with 1D pooling.
We discuss the impact of 2D pooling in Section~\ref{ablation}.
\section{Proposed Method}

In this section, we first briefly revisit the preliminaries of Visual Transformers~\cite{vit} and then introduce our proposed Hierarchical Visual Transformer. 

\subsection{Preliminary}
\label{sec:preliminary}
Let $\mI \in \mathbb{R}^{H \times W \times C}$ be an input image, where $H$, $W$ and $C$ represent the height, width, and the number of channels, respectively. To handle a 2D image, 
ViT first splits the image into a sequence of flattened 2D patches $\bX = [\bx_p^1; \bx_p^2; ...; \bx_p^N]$, where $ \bx_p^i \in \mathbb{R}^{P^{2} C}$ is the $i$-th patch of the input image and $[\cdot]$ is the concatenation operation. Here, $N=HW/P^2$ is the number of patches and $P$ is the size of each patch. 
ViT then uses a trainable linear projection that maps each vectorized patch to a $D$ dimension patch embedding. Similar to the class token in BERT~\cite{Devlin2019BERTPO}, ViT prepends a learnable embedding $\bx_{cls} \in \mathbb{R}^{D}$ to the sequence of patch embeddings.
To retain positional information, ViT introduces an additional learnable positional embeddings $\bE \in \mathbb{R}^{(N+1) \times D}$. Mathematically, the resulting representation of the input sequence can be formulated as
\begin{equation}\label{eq:1}
    \bX_0 =  [\bx_{cls}; \bx_p^1 \bW; \bx_p^2 \bW; ...; \bx_p^N \bW] +  \bE,
\end{equation}
where $ \bW \in  \mathbb{R}^{P^2 C \times D}$ is a learnable linear projection parameter. Then, the resulting sequence of embeddings serves as the input to the Transformer encoder~\cite{transformer}. 

Suppose that the encoder in a Transformer consists of $L$ blocks. Each block contains a multi-head self-attention (MSA) layer and a position-wise multi-layer perceptron (MLP). 
For each layer, layer normalization (LN)~\cite{Ba2016LayerN} and residual connections~\cite{resnet} are employed, 
which can be formulated as follows
\begin{align}
    \label{eq:2}
    \bX^{'}_{l-1} &= \bX_{l-1} + \mathrm{MSA}(\mathrm{LN}(\bX_{l-1})), \\
    \label{eq:3}
    \bX_l &= \bX^{'}_{l-1} + \mathrm{MLP}(\mathrm{LN}(\bX^{'}_{l-1})), 
\end{align}
where $l \in [1, ..., L]$ is the index of Transformer blocks. Here, a $\textrm{MLP}$ contains two fully-connected layers with a GELU non-linearity~\cite{Hendrycks2016GaussianEL}. 
In order to perform classification, ViT applies a layer normalization layer and a fully-connected (FC) layer to the first token of the Transformer encoder's output $\bX_L^0$. In this way, the output prediction $\by$ can be computed by
\begin{equation}\label{eq:4}
    \by = \mathrm{FC}(\mathrm{LN}(\bX_L^0)).
\end{equation}

\subsection{Hierarchical Visual Transformer}
In this paper, we propose a Hierarchical Visual Transformer (HVT) to reduce the redundancy in the full-length patch sequence and construct a hierarchical representation. In the following, we first propose a hierarchical pooling to gradually shrink the sequence length and hence reduce the computational cost. Then, we propose to perform predictions without the class token. The overview of the proposed HVT is shown in Figure~\ref{fig:model_arch}. 

\subsubsection{Hierarchical Pooling}
\label{sec:hierarchical_pooling}
We propose to apply hierarchical pooling in ViT for two reasons: (1) Recent studies~\cite{powerbert,funnel} on Transformers show that tokens tend to carry redundant information as it goes deeper. Therefore, it would be beneficial to reduce these redundancies through the pooling approaches. (2) The input sequence projected from image patches in ViT can be seen as flattened CNN feature maps with encoded spatial information, hence pooling from the nearby tokens can be analogous to the spatial pooling methods in CNNs. 

Motivated by the hierarchical pipeline of VGG-style~\cite{simonyan2014very} and ResNet-style~\cite{resnet} networks, we partition the Transformer blocks into $M$ stages and apply downsampling operation to each stage to shrink the sequence length. 
Let $\{ b_1, b_2, \dots, b_M \}$ be the indexes of the first block in each stage. At the $m$-th stage, we apply a 1D max pooling operation with a kernel size of $k$ and stride of $s$ to the output of the Transformer block $b_m \in \{ b_1, b_2, \dots, b_M \}$ to shrink the sequence length.

Note that the positional encoding is important for a Transformer since the positional encoding is able to capture information about the relative and absolute position of the token in the sequence~\cite{transformer,detr}.
In Eq.~(\ref{eq:1}) of ViT, each patch is equipped with positional embedding $\bE$ at the beginning. However, in our HVT, the original positional embedding $\bE$ may no longer be meaningful after pooling since the sequence length is reduced after each pooling operation. 
In this case, positional embedding in the pooled sequence needs to be updated. 
Moreover, previous work~\cite{funnel} in NLP also find it important to complement positional information after changing the sequence length. Therefore, at the $m$-th stage, we introduce an additional learnable positional embedding $\bE_{b_m}$ to capture the positional information, which can be formulated as

\begin{equation} \label{eq:5}
    \hat{\bX}_{b_m} = \mathrm{MaxPool1D}(\bX_{b_m}) +  \bE_{b_m},
\end{equation}
where $\bX_{b_m}$ is the output of the Transformer block $b_m$. We then forward the resulting embeddings $\hat{\bX}_{b_m}$ into the next Transformer block $b_m + 1$.

\subsubsection{Prediction without the Class Token}
Previous works \cite{vit, deit} make predictions by taking the class token as input in classification tasks as described in Eq.~(\ref{eq:4}). However, such structure relies solely on the single class token with limited capacity while discarding the remaining sequence that is capable of storing more discriminative information. To this end, we propose to remove the class token in the first place and predict with the remaining output sequence on the last stage. 

Specifically, given the output sequence without the class token on the last stage $\bX_L$, we first apply average pooling, then directly apply an FC layer on the top of the pooled embeddings and make predictions. The process can be formulated as
\begin{equation}\label{eq:6}
    \by = \mathrm{FC}(\mathrm{AvgPool}(\mathrm{LN}(\bX_{L}))).
\end{equation}

\begin{figure*}[htp!]
	\centering
	\includegraphics[width=0.98\textwidth]{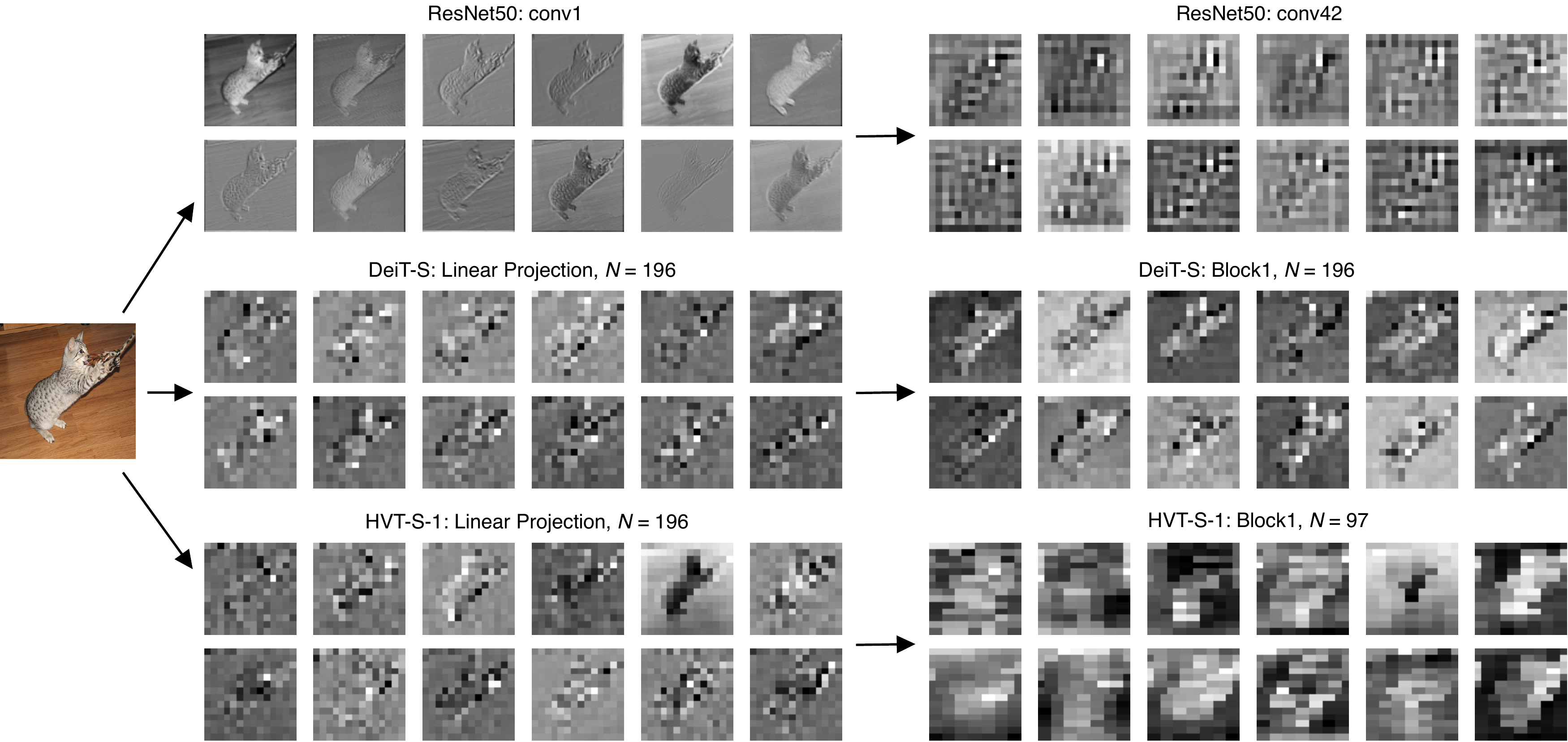}
	\caption{Feature visualization of ResNet50~\cite{resnet}, DeiT-S~\cite{deit} and our HVT-S-1 trained on ImageNet. DeiT-S and our HVT-S-1 correspond to the small setting in DeiT, except that our model applies a pooling operation and performing predictions without the class token. The resolution of the feature maps from ResNet50 conv1 and conv4\_2 are 112$\times$112 and 14$\times$14, respectively. For DeiT and HVT, the feature maps are reshaped from tokens. For our model, we interpolate the pooled sequence to its initial length then reshape it to a 2D map.
	}
	\label{fig:visualization}
	\vspace{-10pt}
\end{figure*}

\subsection{Complexity Analysis}
In this section, we analyse the block-wise compression ratio with hierarchical pooling.
Following ViT~\cite{vit}, we use FLOPs to measure the computational cost of a Transformer. Let $n$ be the number of tokens in a sequence and $d$ is the dimension of each token. The FLOPs of a Transformer block $\phi_{BLK}(n,d)$ can be computed by
\begin{equation}\label{eq:blk_flops}
\begin{split}
   \phi_{BLK}(n,d) &= \phi_{MSA}(n,d) + \phi_{MLP}(n,d), \\
                 &= 12nd^2 + 2n^2d,
\end{split}
\end{equation}
where $\phi_{MSA}(n,d)$ and $\phi_{MLP}(n,d)$ are the FLOPs of the MSA and MLP, respectively. Details about Eq.~(\ref{eq:blk_flops}) can be found in the supplementary material.

Without loss of generality, suppose that the sequence length $n$ is reduced by half after performing hierarchical pooling. In this case, the block-wise compression ratio $\alpha$ can be computed by
\begin{equation}\label{eq:blk_com_ratio}
    \alpha = \frac{\phi_{BLK}(n,d)}{\phi_{BLK}(n/2,d)} = 2 + \frac{2}{12(d/n) + 1}.
\end{equation}

Clearly, Eq.~(\ref{eq:blk_com_ratio}) is monotonic, thus the block-wise compression ratio $\alpha$ is bounded by $(2, 4)$, \ie, $\alpha \in (2, 4)$. 

\section{Discussions}

\subsection{Analysis of Hierarchical Pooling}

In CNNs, feature maps are usually downsampled to smaller sizes in a hierarchical way \cite{simonyan2014very,resnet}. In this paper, we show that this principle can be applied to ViT models by comparing the visualized feature maps from ResNet conv4\_2, DeiT-S~\cite{deit} block1 and HVT-S-1 block1 in Figure~\ref{fig:visualization}. From the figure, in ResNet, the initial feature maps after the first convolutional layer contain rich edge information. After feeding the features to consecutive convolutional layers and a pooling layer, the output feature maps tend to preserve more high-level discriminative information. In DeiT-S, following the ViT structure, although the image resolution for the feature maps has been reduced to 14 $\times$ 14 by the initial linear projection layer, we can still observe clear edges and patterns. Then, the features get refined in the first block to obtain sharper edge information. In contrast to DeiT-S that refines features at the same resolution level, after the first block, the proposed HVT downsamples the hidden sequence through a pooling layer and reduces the sequence length by half. We then interpolate the sequence back to 196 and reshape it to 2D feature maps. We can find that the hidden representations contain more abstract information with high discriminative power, which is similar to ResNet.

\subsection{Scalability of HVT}
The computational complexity reduction equips HVT with strong scalability in terms of width/depth/patch size/resolution. Take DeiT-S for an example, the model consists of 12 blocks and 6 heads. Given a 224$\times$224 image with a patch size of 16, the computational cost of DeiT-S is around 4.6G FLOPs. By applying four pooling operations, our method is able to achieve nearly 3.3$\times$ FLOPs reduction. Furthermore, to re-allocate the reduced FLOPs, we may construct wider or deeper HVT-S, with 11 heads or 48 blocks, then the overall FLOPs would be around 4.51G and 4.33G, respectively. Moreover, we may consider a longer sequence by setting a smaller patch size or using a larger resolution. For example, with a patch size of 8 and an image resolution of 192$\times$192, the FLOPs for HVT-S is around 4.35G. Alternatively, enlarging the image resolution into 384$\times$384 will lead to 4.48G FLOPs. In all of the above mentioned cases, the computational costs are still lower than that of DeiT-S while the model capacity is enhanced. 

It is worth noting that finding a principled way to scale up HVT to obtain the optimal efficiency-vs-accuracy tradeoff remains an open question. At the current stage, we take an early exploration by evenly partitioning blocks and following model settings in DeiT~\cite{deit} for a fair comparison. In fact, the improved scalability of HVT makes it possible for using Neural Architecture Search (NAS) to automatically find optimal configurations, such as EfficientNet~\cite{tan2019efficientnet}. We leave for more potential studies for future work.

\begin{table*}[]
\renewcommand\arraystretch{1.3}
\centering
\caption{
Performance comparisons with DeiT and PoWER on ImageNet. ``Embedding Dim'' refers to the dimension of each token in the sequence. ``\#Heads'' and ``\#Blocks'' are the number of self-attention heads and blocks in Transformer, respectively. ``FLOPs'' is measured with a $224 \times 224$ image. 
``Ti'' and ``S'' are short for the tiny and small settings, respectively. “Architecture-$M$” denotes the model with $M$ pooling stages. ``Scale'' denotes that we scale up the embedding dimension and/or the number of self-attention heads. ``DeiT-Ti/S + PoWER'' refers to the model that applies the techniques in PoWER-BERT~\cite{powerbert} to DeiT-Ti/S.
}
\resizebox{0.9\textwidth}{!} {
\fontsize{11}{11}\selectfont
\begin{tabular}{@{}lccccc|ll@{}}
Model & Embedding Dim & \#Heads & \#Blocks & FLOPs (G) & Params (M) & Top-1 Acc. (\%) & Top-5 Acc. (\%) \\ 
\shline
DeiT-Ti~\cite{deit} & 192 & 3 & 12 & 1.25 & 5.72 & 72.20 & 91.10 \\
DeiT-Ti + PoWER~\cite{powerbert}  & 192 & 3 & 12 & 0.80 & 5.72 & 69.40 {\color[HTML]{CB0000} \fontsize{9pt}{9pt}\selectfont{(-2.80)}} & 89.20 {\color[HTML]{CB0000}\fontsize{9pt}{9pt}\selectfont{(-1.90)}} \\
HVT-Ti-1 & 192 & 3 & 12 & 0.64 & 5.74 & 69.64 {\color[HTML]{CB0000}\fontsize{9pt}{9pt}\selectfont{(-2.56)}} & 89.40 {\color[HTML]{CB0000}\fontsize{9pt}{9pt}\selectfont{(-1.70)}} \\ 
Scale HVT-Ti-4 & 384 & 6 & 12 & 1.39 & 22.12 & 75.23 {\color[HTML]{009901} \fontsize{9pt}{9pt}\selectfont{(+3.03)}} & 92.30 {\color[HTML]{009901}  \fontsize{9pt}{9pt}\selectfont{(+1.20)}} \\ \hline
DeiT-S~\cite{deit} & 384 & 6 & 12 & 4.60 & 22.05 & 79.80 & 95.00 \\
DeiT-S + PoWER~\cite{powerbert}  & 384 & 6 & 12 & 2.70 & 22.05 & 78.30  {\color[HTML]{CB0000}\fontsize{9pt}{9pt}\selectfont{(-1.50)}}  & 94.00  {\color[HTML]{CB0000}\fontsize{9pt}{9pt}\selectfont{(-1.00)}} \\
HVT-S-1 & 384 & 6 & 12 & 2.40 & 22.09 & 78.00 {\color[HTML]{CB0000}\fontsize{9pt}{9pt}\selectfont{(-1.80)}} & 93.83  {\color[HTML]{CB0000}\fontsize{9pt}{9pt}\selectfont{(-1.17)}}
\end{tabular}
}
\label{tab:compare_imagenet}
\vspace{-5pt}
\end{table*}

\begin{table*}[]
\renewcommand\arraystretch{1.2}
\centering
\caption{Effect of the prediction without the class token. ``CLS'' denotes the class token. 
}
\vspace{-5pt}
\resizebox{0.9\textwidth}{!} {
\fontsize{13}{13}\selectfont
\begin{tabular}{lcc|ll|ll}
\multirow{2}{*}{Model} & \multirow{2}{*}{FLOPs (G)} & \multirow{2}{*}{Params (M)} & \multicolumn{2}{c}{ImageNet} & \multicolumn{2}{c}{CIFAR-100} \\ \cline{4-7} 
 &  &  & Top-1 Acc. (\%)  & Top-5 Acc. (\%) & Top-1 Acc. (\%)  & Top-5 Acc. (\%) \\ \shline
DeiT-Ti with CLS & 1.25 & 5.72 & 72.20 & 91.10 & 64.49 & 89.27 \\
DeiT-Ti without CLS & 1.25 & 5.72 & 72.42{\color[HTML]{009901} \fontsize{9pt}{9pt}\selectfont{ (+0.22)}} & 91.55{\color[HTML]{009901} \fontsize{9pt}{9pt}\selectfont{ (+0.45)}} & 65.93{\color[HTML]{009901} \fontsize{9pt}{9pt}\selectfont{ (+1.44)}} & 90.33{\color[HTML]{009901} \fontsize{9pt}{9pt}\selectfont{ (+1.06)}} \\
\end{tabular}
}
\label{tab:cls_token}
\vspace{-5pt}
\end{table*}

\section{Experiments} \label{experiment}

\paragraph{Compared methods.} To investigate the effectiveness of HVT, we 
compare our method with DeiT~\cite{deit} and a BERT-based pruning method PoWER-BERT~\cite{powerbert}. DeiT is a representative Vision Transformer and PoWER progressively prunes unimportant tokens in pretrained BERT models for inference acceleration. Moreover, we 
consider two architectures in DeiT for comparisons: \textbf{HVT-Ti}: HVT with the tiny setting. \textbf{HVT-S}: HVT with the small setting. For convenience, we use “Architecture-$M$” to represent our model with $M$ pooling stages, \eg, HVT-S-1.

\paragraph{Datasets and Evaluation metrics.} We evaluate our proposed HVT on two image classification benchmark datasets: CIFAR-100 \cite{krizhevsky2009learning} and ImageNet \cite{russakovsky2015imagenet}. We measure the performance of different methods in terms of the Top-1 and Top-5 accuracy. Following DeiT~\cite{deit}, we measure the computational cost by FLOPs. Moreover, we also measure the model size by the number of parameters (Params).

\paragraph{Implementation details.} 
For experiments on ImageNet, we train our models for 300 epochs with a total batch size of 1024. The initial learning rate is 0.0005. We use AdamW optimizer \cite{loshchilov2019decoupled} with a momentum of 0.9 for optimization. We set the weight decay to 0.025. For fair comparisons, we keep the same data augmentation strategy as DeiT~\cite{deit}. For the downsampling operation, we use max pooling by default. The kernel size $k$ and stride $s$ are set to 3 and 2, respectively, chosen by a simple grid search on CIFAR100. Besides, all learnable positional embeddings are initialized in the same way as DeiT. 
More detailed settings on the other hyper-parameters can be found in DeiT. For experiments on CIFAR-100, we train our models with a total batch size of 128. The initial learning rate is set to 0.000125. Other hyper-parameters are kept the same as those on ImageNet.
\begin{figure}[!tp]
	\centering
	\includegraphics[width=0.80\linewidth]{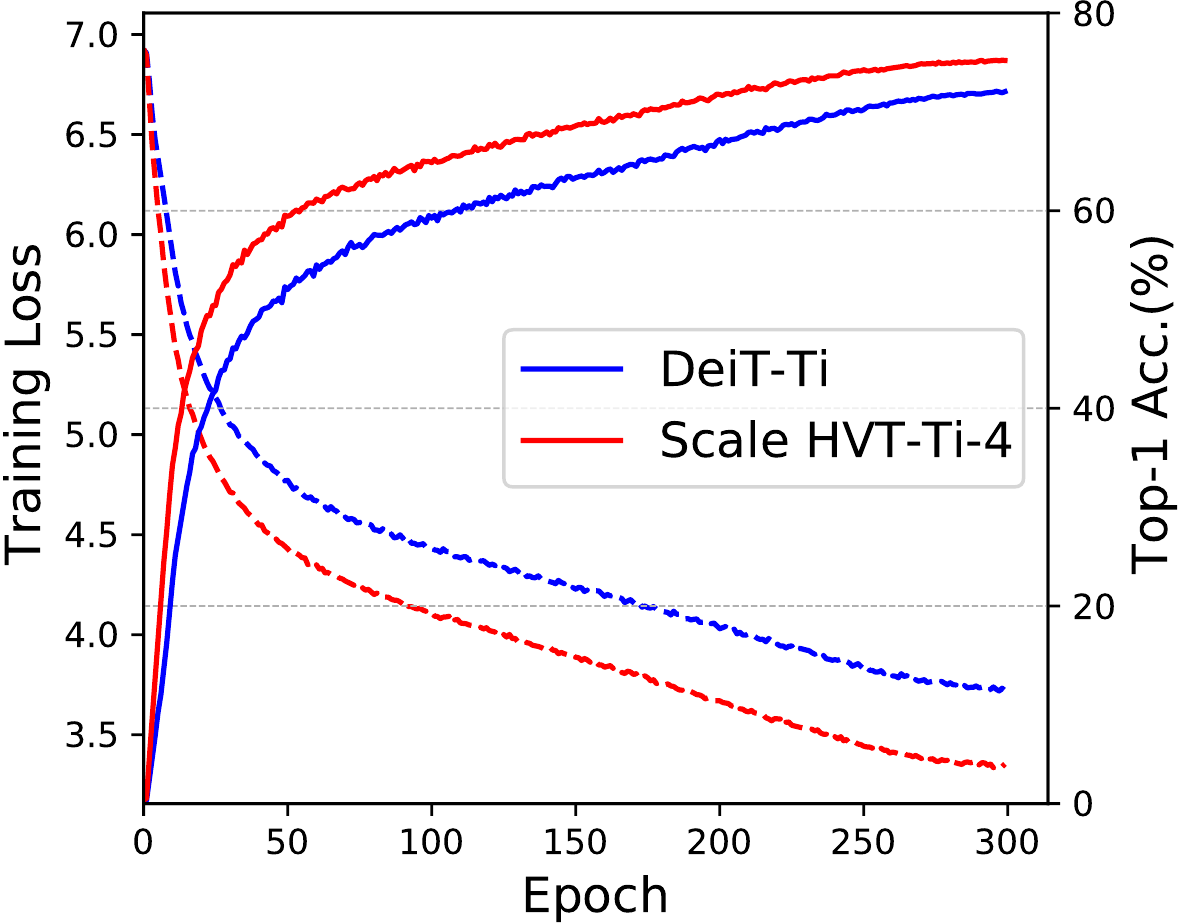}
	\vspace{-5pt}
	\caption{Performance comparisons of DeiT-Ti (1.25G FLOPs) and the proposed Scale HVT-Ti-4 (1.39G FLOPs). All the models are evaluated on ImageNet.
	Solid lines denote the Top-1 accuracy (y-axis on the right). Dash lines denote the training loss (y-axis on the left). 
	}
	\label{fig:same_flops_compare}
	\vspace{-18pt}
\end{figure}
\subsection{Main Results} 
We compare the proposed HVT with DeiT and PoWER, and report the results in Table~\ref{tab:compare_imagenet}. First, compared to DeiT, our HVT achieves nearly 2$\times$ FLOPs reduction with a hierarchical pooling.
However, the significant FLOPs reduction also leads to performance degradation in both the tiny and small settings. Additionally, the performance drop of HVT-S-1 is smaller than that of HVT-Ti-1. For example, for HVT-S-1, it only leads to $1.80\%$ drop in the Top-1 accuracy. In contrast, it results in $2.56\%$ drop in the Top-1 accuracy for HVT-Ti-1. It can be attributed to that, compared with HVT-Ti-1, HVT-S-1 is more redundant with more parameters. Therefore, applying hierarchical pooling to HVT-S-1 can significantly reduce redundancy while maintaining performance. Second, compared to PoWER, HVT-Ti-1 uses less FLOPs while achieving better performance. Besides, HVT-S-1 reduces more FLOPs than PoWER, while achieving slightly lower performance than PoWER. Also note that PoWER involves three training steps, while ours is a simpler one-stage training scheme.

Moreover, we also compare the scaled HVT with DeiT under similar FLOPs. Specifically, we enlarge the embedding dimensions and add extra heads in HVT-Ti. From Table~\ref{tab:compare_imagenet} and Figure~\ref{fig:same_flops_compare}, by re-allocating the saved FLOPs to scale up the model, HVT can converge to a better solution and yield improved performance. For example, the Top-1 accuracy on ImageNet can be improved considerably by 3.03\% in the tiny setting. More empirical studies on the effect of model scaling can be found in Section~\ref{ablation}.

\subsection{Ablation Study} 
\label{ablation}

\paragraph{Effect of the prediction without the class token.}
To investigate the effect of the prediction without the class token, we train DeiT-Ti with and without the class token and show the results in Table~\ref{tab:cls_token}.  
From the results, the models without the class token outperform the ones with the class token. The performance gains mainly come from the extra  discriminative information stored in the entire sequence without the class token.
Note that the performance improvement on CIFAR-100 is much larger than that on ImageNet. It may be attributed that CIFAR-100 is a small dataset, which lacks varieties compared with ImageNet. Therefore, the model trained on CIFAR-100 benefits more from the increase of model's discriminative power.

\begin{table}[]
\centering
\caption{Performance comparisons on HVT-S-4 with three downsampling operations: convolution, max pooling and average pooling. We report the Top-1 and Top-5 accuracy on CIFAR-100.
}
\vspace{-5pt}
\renewcommand\arraystretch{1.2}
\resizebox{\columnwidth}{!}{
\fontsize{15}{15}\selectfont
\begin{tabular}{@{}cccc|cc@{}}
Model & Operation & FLOPs (G) & Params (M) & Top-1 Acc. (\%)  & Top-5 Acc. (\%) \\ \shline
HVT-S & Conv       & 1.47     & 23.54     & 69.75          & 92.12 \\
HVT-S & Avg       & 1.39     & 21.77     & 70.38          & 91.39          \\
HVT-S & Max       & 1.39     & 21.77     & 75.43          & 93.56        \\ 
\end{tabular}
}
\label{tab:pool_type}
\vspace{-10pt}
\end{table}

\begin{table}[]
\caption{Performance comparisons on HVT-S with different pooling stages $M$. We report the Top-1 and Top-5 accuracy on CIFAR-100.
}
\vspace{-10pt}
\renewcommand\arraystretch{1.1}
\resizebox{\columnwidth}{!}{
\begin{tabular}{ccc|cccc}
\multirow{2}{*}{$M$} & \multirow{2}{*}{FLOPs} & \multirow{2}{*}{Params} & \multicolumn{2}{c}{ImageNet} & \multicolumn{2}{c}{CIFAR100} \\ \cline{4-7} 
  &      &       & Top-1 (\%) & \multicolumn{1}{c|}{Top-5 (\%)} & Top-1 (\%) & Top-5 (\%) \\ \hline
0 & 4.57 & 21.70 & 80.39           & \multicolumn{1}{c|}{95.13}            & 71.99       & 92.44       \\
1 & 2.40 & 21.74 & 78.00   & \multicolumn{1}{c|}{93.83}       & 74.27       & 93.07       \\
2 & 1.94 & 21.76 & 77.36       & \multicolumn{1}{c|}{93.55}       & 75.37       & 93.69       \\
3 & 1.62 & 21.77 & 76.32       & \multicolumn{1}{c|}{92.90}        & 75.22       & 93.90       \\
4 & 1.39 & 21.77 & 75.23       & \multicolumn{1}{c|}{92.30}        & 75.43       & 93.56      
\end{tabular}
}
\label{tab:pooling_times}
\vspace{-10pt}
\end{table}

\paragraph{Effect of different pooling stages.}
We train HVT-S with different pooling stages $M \in \{ 0, 1, 2, 3, 4\}$ and show the results in Table~\ref{tab:pooling_times}. Note that HVT-S-0 is equivalent to the DeiT-S without the class token.
With the increase of $M$, HVT-S achieves better performance with decreasing FLOPs on CIFAR-100, while on ImageNet we observe the accuracy degrades.
One possible reason is that HVT-S is very redundant on CIFAR-100, such that pooling acts as a regularizer to avoid the overfitting problem and improves the generalization of HVT on CIFAR-100. On ImageNet, we assume HVT is less redundant and a better scaling strategy is required to improve the performance.

\paragraph{Effect of different downsampling operations.}
To investigate the effect of different downsampling operations, we train HVT-S-4 with three downsampling strategies: convolution, average pooling and max pooling. 
As Table~\ref{tab:pool_type} shows, downsampling with convolution performs the worst even it introduces additional FLOPs and parameters. Besides, average pooling performs slightly better than convolution in terms of the Top-1 accuracy. Compared with the two settings, HVT-S-4 with max pooling performs much better as it significantly surpasses average pooling by 5.05\% in the Top-1 accuracy and 2.17\% in the Top-5 accuracy. The result is consistent with the common sense~\cite{pool_anly} that max pooling performs well in a large variety of settings. To this end, we use max pooling in all other experiments by default.


\begin{table}[]
\caption{
Performance comparisons on HVT-S-4 with different number of Transformer blocks. We report the Top-1 and Top-5 accuracy on CIFAR-100.
}
\vspace{-3pt}
\renewcommand\arraystretch{1.1}
\resizebox{\columnwidth}{!} {
\begin{tabular}{@{}ccc|cc@{}}
\#Blocks & FLOPs (G) & Params (M) & Top-1 Acc. (\%)  & Top-5 Acc. (\%) \\ \shline
12 & 1.39 & 21.77 & 75.43 & 93.56 \\ 
16 & 1.72 & 28.87 & 75.32 & 93.30 \\
20 & 2.05 & 35.97 & 75.35 & 93.35 \\
24 & 2.37 & 43.07 & 75.04 & 93.39 \\
\end{tabular}
}
\label{tab:num_blocks}
\vspace{-8pt}
\end{table}

\begin{table}[]
\centering
\caption{
Performance comparisons on HVT-Ti-4 with different number of self-attention heads. We report the Top-1 and Top-5 accuracy on CIFAR-100.
}
\vspace{-5pt}
\renewcommand\arraystretch{1.1}
\resizebox{\columnwidth}{!} {
\begin{tabular}{@{}ccc|cc@{}}
\#Heads & FLOPs (G) & Params (M) & Top-1 Acc. (\%)  & Top-5 Acc. (\%) \\ \shline
3 & 0.38 & 5.58 & 69.51 & 91.78 \\
6 & 1.39 & 21.77 & 75.43 & 93.56 \\
12 & 5.34 & 86.01 & 76.26 & 93.39 \\ 
16 & 9.39 & 152.43 & 76.30 & 93.16 \\ 
\end{tabular}
}
\label{tab:num_heads}
\vspace{-15pt}
\end{table}

\paragraph{Effect of model scaling.}
\label{sec:model_scaling}
One of the important advantages of the proposed hierarchical pooling is that we can re-allocate the saved computational cost for better model capacity by constructing a model with a wider, deeper, larger resolution or smaller patch size configuration.
Similar to the CNNs literature \cite{resnet,wider_deeper,wrn}, we study the effect of model scaling in the following.

Based on HVT-S-4, we first construct deeper models by increasing the number of blocks in Transformers. Specifically, we train 4 models with different number of blocks $L \in \{12, 16, 20, 24\}$. As a result, each pooling stage for different models would have 3, 4, 5, and 6 blocks, respectively. We train 4 models on CIFAR-100 and report the results in Table~\ref{tab:num_blocks}. From the results, we observe no more gains by stacking more blocks in HVT. 

Based on HVT-Ti-4, we then construct wider models by increasing the number of self-attention heads. To be specific, we train 4 models with different numbers of self-attention heads, \ie, 3, 6, 12, and 16, on CIFAR-100 and report the results in Table~\ref{tab:num_heads}. From the results, our models achieve better performance with the increase of width. For example, the model with 16 self-attention heads outperforms those with 3 self-attention heads by 6.79\% in the Top-1 accuracy and 1.38\% in the Top-5 accuracy.

Based on HVT-S-4, we further construct models with larger input image resolutions. Specifically, we train 4 models with different input image resolutions, \ie, 160, 224, 320, and 384, on CIFAR-100 and report the results in Table~\ref{tab:resolution}. From the results, with the increase of image resolution, our models achieve better performance. For example, the model with the resolution of 384 outperforms those with the resolution of 160 by 2.47\% in the Top-1 accuracy and 1.12\% in the Top-5 accuracy. Nevertheless, increasing image resolutions also leads to high computational cost. To make a trade-off between computational cost and accuracy, we set the image resolution to 224 by default.

We finally train HVT-S-4 with different patch sizes $P \in \{ 8, 16, 32 \}$ and show the results in Table~\ref{tab:patch_size}. From the results, HVT-S-4 performs better with the decrease of patch size. 
For example, when the patch size decreases from 32 to 8, our HVT-S achieves 9.14\% and 4.03\% gain in terms of the Top-1 and Top-5 accuracy. Intuitively, a smaller patch size leads to fine-grained image patches and helps to learn high-resolution representations, which is able to improve the classification performance. However, with a smaller patch size, the patch sequence will be longer, which significantly increases the computational cost. To make a balance between the computational cost and accuracy, we set the patch size to 16 by default.

\paragraph{Exploration on 2D pooling.}

Compared to 1D pooling, 2D pooling brings more requirements. For example, it requires a smaller patch size to ensure a sufficient sequence length. Correspondingly, it is essential to reduce the heads at the early stages to save FLOPs and memory consumption from high-resolution feature maps. Besides, it also requires to vary the blocks at each stage to control the overall model complexity. In Table~\ref{tab:2d_pooling}, we apply 2D pooling to HVT-S-2 and compare it with DeiT-S. The results show that HVT-S-2 with 2D pooling outperforms DeiT-S on CIFAR100 by a large margin with similar FLOPs. In this case, we assume that HVT can achieve promising performance with a dedicated scaling scheme for 2D pooling. We will leave this exploration for future work.

\begin{table}[]
\caption{Performance comparisons on HVT-S-4 with different image resolutions. We report the Top-1 and Top-5 accuracy on CIFAR-100.
}
\vspace{-5pt}
\renewcommand\arraystretch{1.1}
\resizebox{\columnwidth}{!} {
\begin{tabular}{ccc|cc}
Resolution & FLOPs (G) & Params (M) & Top-1 Acc. (\%)  & Top-5 Acc. (\%) \\ \shline
160 & 0.69 & 21.70 & 73.84 & 92.90 \\
224 & 1.39 & 21.77 & 75.43 & 93.56 \\
320 & 3.00 & 21.92 & 75.54 & 94.18 \\
384 & 4.48 & 22.06 & 76.31 & 94.02 \\ 
\end{tabular}
}
\label{tab:resolution}
\vspace{-12pt}
\end{table}

\begin{table}[]
\centering
\caption{Performance comparisons on HVT-S-4 with different patch sizes $P$. We report the Top-1 and Top-5 accuracy on CIFAR-100.
}
\vspace{-5pt}
\renewcommand\arraystretch{1.1}
\resizebox{\columnwidth}{!} {
\begin{tabular}{@{}ccc|cc@{}}
$P$ & FLOPs (G) & Params (M) & Top-1 Acc. (\%)  & Top-5 Acc. (\%) \\ \shline
8 & 6.18  & 21.99 & 77.29  & 94.22  \\ 
16 & 1.39 & 21.77 & 75.43  & 93.56 \\ 
32 & 0.37 & 22.55 & 68.15 & 90.19 \\
\end{tabular}
}
\label{tab:patch_size}
\vspace{-10pt}
\end{table}

\begin{table}[]
\centering
\caption{Effect of 2D pooling on HVT-S-2.  We report the Top-1 and Top-5 accuracy on CIFAR-100. For HVT-S-2, we apply 2D max pooling and use a patch size of 8.}
\vspace{-5pt}
\renewcommand\arraystretch{1.1}
\resizebox{\columnwidth}{!} {
\begin{tabular}{l|cccc}
Model        & FLOPs (G) & Params (M) & Top-1 Acc. (\%) & Top-5 Acc. (\%) \\ \shline
DeiT-S       & 4.60       & 21.70      & 71.99           & 92.44           \\
HVT-S-2 (2D) & 4.62      & 21.80      & 77.58           & 94.40           
\end{tabular}
}
\label{tab:2d_pooling}
\vspace{-15pt}
\end{table}


\section{Conclusion and Future Work}

In this paper, we have presented a Hierarchical Visual Transformer, termed HVT, for image classification.
In particular, the proposed hierarchical pooling can significantly compress the sequential resolution to save computational cost in a simple yet effective form. 
More importantly, this strategy greatly improves the scalability of visual Transformers, making it possible to scale various dimensions - depth, width, resolution and patch size. By re-allocating the saved computational cost, we can scale up these dimensions for better model capacity with comparable or fewer FLOPs.
Moreover, we have empirically shown that the visual tokens are more important than the single class token for class prediction. 
Note that the scope of this paper only targets designing our HVT as an encoder. 
Future works may include extending our HVT model to decoder and to solve other mainstream CV tasks, such as object detection and semantic/instance segmentation. In addition, it would be interesting to find a principled way to scale up HVT that can achieve better accuracy and efficiency. 
\section{Acknowledgements}
This research is partially supported by Monash FIT Start-up Grant and Sensetime Gift Fund.

{\small
\bibliographystyle{ieee_fullname}
\bibliography{egbib}
}

\setcounter{section}{0}

\renewcommand{\thesection}{S\arabic{section}}

\section*{Appendix}

\renewcommand\thefigure{S\arabic{figure}}
\renewcommand{\thetable}{S\arabic{table}}

We organize our supplementary material as follows. 
\begin{itemize}
    \item In Section~\ref{transformer_block}, we elaborate on the components of a Transformer block, including the multi-head self-attention layer (MSA) and the position-wise multi-layer perceptron (MLP).
    \item In Section~\ref{block_flops}, we provide details for the FLOPs calculation of a Transformer block.
\end{itemize}

\section{Transformer Block} \label{transformer_block}
\subsection{Multi-head Self-Attention}
Let $\bX \in \mathbb{R}^{N \times D} $ be the input sentence, where $N$ is the sequence length and $D$ the embedding dimension. First, a self-attention layer computes query, key and value matrices from $\bX$ using linear transformations
\begin{equation}
  [\bQ, \bK, \bV] = \bX \bW_{qkv},
\end{equation}
where $\bW_{qkv} \in  \mathbb{R}^{D \times 3D_{h}}$ is a learnable parameter and $D_h$ is the dimension of each self-attention head.
Next, the attention map $\bA$ can be calculated by scaled inner product from $\bQ$ and $\bK$ and normalized by a softmax function
\begin{equation}
  \bA = \mathrm{Softmax}(\bQ\bK^\top / \sqrt{D_h}),
\end{equation}
where $\bA \in \mathbb{R}^{N\times N}$ and $A_{ij}$ represents for the attention score between the $\bQ_{i}$ and $\bK_{j}$. Then, the self-attention operation is applied on the value vectors to produce an output matrix
\begin{equation}
    \bO =\bA\bV,
\end{equation}
where $\bO \in \mathbb{R}^{N\times D_h}$. For a multi-head self-attention layer with $D/D_h$ heads, the outputs can be calculated by a linear projection for the concatenated self-attention outputs
\begin{equation}
    \bX^{'} = [\bO_1;\bO_2; ...; \bO_{D/D_h}]\bW_{proj},
\end{equation}
where $\bW_{proj} \in  \mathbb{R}^{D \times D}$ is a learnable parameter and $[\cdot]$ denotes the concatenation operation.

\subsection{Position-wise Multi-Layer Perceptron}
Let $\bX^{'}$ be the output from the MSA layer. An MLP layer which contains two fully-connected layers with a GELU non-linearity can be represented by
\begin{equation}
    \bX = \mathrm{GELU}(\bX^{'} \bW_{fc1}) \bW_{fc2},
\end{equation}
where $\bW_{fc1} \in \mathbb{R}^{D \times 4D}$ and $\bW_{fc2} \in \mathbb{R}^{4D \times D}$ are learnable parameters.

\section{FLOPs of a Transformer Block} \label{block_flops}
We denote $\phi(n, d)$ as a function of FLOPs with respect to the sequence length $n$ and the embedding dimension $d$. 
For an MSA layer, The FLOPs mainly comes from four parts: (1) The projection of $\bQ$,$\bK$,$\bV$ matrices $\phi_{qkv}(n, d) = 3nd^2$. (2) The calculation of the attention map $\phi_{A}(n, d) = n^2d$. (3) The self-attention operation $\phi_{O}(n, d) =n^2d$. (4) And finally, a linear projection for the concatenated self-attention outputs $\phi_{proj}(n, d)=nd^2$. Therefore, the overall FLOPs for an MSA layer is
\begin{equation} \label{eq:msa_flops}
    \begin{split}
         \phi_{MSA}(n, d) &= \phi_{qkv}(n, d) + \phi_{A}(n, d) + \phi_{O}(n, d) + \phi_{proj}(n, d) \\
         &= 3nd^2 + n^2d + n^2d + nd^2 \\
         &= 4nd^2 + 2n^2d.
    \end{split}
\end{equation}

For an MLP layer, the FLOPs mainly comes from two fully-connected (FC) layers. The first FC layer $fc1$ is used to project each token from $\mathbb{R}^{d}$ to $\mathbb{R}^{4d}$. The next FC layer $fc2$  projects each token back to $\mathbb{R}^{d}$. Therefore, the FLOPs for an MLP layer is
\begin{equation} \label{eq:mlp_flops}
    \phi_{MLP}(n, d) = \phi_{fc1}(n, d) + \phi_{fc2}(n, d) = 4nd^2 + 4nd^2 = 8nd^2.
\end{equation}

By combining Eq.~(\ref{eq:msa_flops}) and Eq.~(\ref{eq:mlp_flops}), we can get the total FLOPs of one Transformer block
\begin{equation}
    \phi_{BLK}(n, d) = \phi_{MSA}(n, d) + \phi_{MLP}(n, d) = 12nd^2 + 2n^2d.
\end{equation}

\end{document}